\documentclass[trsc,nonblindrev]{informs3} 

\OneAndAHalfSpacedXI 

\renewcommand{\theARTICLETOP}{}
\renewcommand{\theRRHFirstLine}{}
\renewcommand{\theRRHSecondLine}{}
\renewcommand{\theLRHFirstLine}{}
\renewcommand{\theLRHSecondLine}{}
\usepackage{soul}
\usepackage{natbib}
 \bibpunct[, ]{(}{)}{,}{a}{}{,}%
 \def\bibfont{\small}%
 \usepackage{mathtools}
\usepackage{mathpazo}
\usepackage{amsmath}
\usepackage{mathrsfs}
\usepackage{amssymb}
\usepackage{multirow}
\usepackage{longtable}
\usepackage{graphicx}
\usepackage{enumerate}
\usepackage{graphicx} 
\usepackage{epstopdf}
\usepackage{svg}
\usepackage{subfigure}
\usepackage{algorithm}
\usepackage{algorithmic}
\usepackage{longtable}
\usepackage{graphicx}
\usepackage{indentfirst}
\usepackage[colorlinks=true, linkcolor=blue, citecolor=blue, urlcolor=blue]{hyperref}

\usepackage{booktabs}
\usepackage{array, caption, threeparttable,arydshln}
\usepackage{caption}
\DeclareCaptionLabelSeparator{twospace}{\ ~\ ~}   
\usepackage{empheq}
\usepackage{cases}

\allowdisplaybreaks[1]
\usepackage{longtable}
\usepackage{multicol} 

\makeatletter
\newcommand{\mathleft}{\@fleqntrue\@mathmargin\parindent}
\newcommand{\mathcenter}{\@fleqnfalse}
\makeatother
\usepackage{etoolbox}
\newcommand{\zerodisplayskips}{%
\setlength{\abovedisplayskip}{7pt}%
\setlength{\belowdisplayskip}{6pt}%
\setlength{\abovedisplayshortskip}{7pt}%
}
\appto{\normalsize}{\zerodisplayskips}
\appto{\small}{\zerodisplayskips}
\appto{\footnotesize}{\zerodisplayskips}

\allowdisplaybreaks \allowdisplaybreaks[1]

\EquationsNumberedThrough    

\MANUSCRIPTNO{} 

\begin{document}

\TITLE{Developing Fundamental Diagrams for Urban Air Mobility Traffic Based on Physical Experiments}

\ARTICLEAUTHORS{%

\AUTHOR{Hang Zhou$^1$, Yuhui Zhai$^2$, Shiyu Shen$^2$, Yanfeng Ouyang$^{2*}$, Xiaowei Shi$^3$, and Xiaopeng Li$^{1*}$}
\AFF{$^1$Department of Civil and Environmental Engineering, University of Wisconsin-Madison, Madison, WI, 53706, USA}
\AFF{$^2$Department of Civil and Environmental Engineering, University of Illinois Urbana-Champaign, Urbana, IL, 61801, USA}
\AFF{$^3$Department of Civil and Environmental Engineering, University of Wisconsin-Milwaukee, Milwaukee, WI, 53211, USA}
\AFF{$^*$Corresponding authors. Email: yfouyang@illinois.edu (Yanfeng Ouyang), xli2485@wisc.edu (Xiaopeng Li)}

}

\ABSTRACT{
    Urban Air Mobility (UAM) is an emerging application of unmanned aerial vehicles that promises to reduce travel time and alleviate congestion in urban transportation systems. As drone density increases, UAM traffic is expected to experience congestion similar to that in ground traffic. However, the fundamental characteristics of UAM traffic, particularly under real-world operating conditions, remain largely unexplored. This study proposes a general framework for constructing the fundamental diagram (FD) of UAM traffic by integrating theoretical analysis with physical experiments. To the best of our knowledge, this is the first study to derive UAM FDs using real-world physical experiment data. On the theoretical side, we design two drone control laws for collision avoidance and develop simulation-based traffic generation methods to produce diverse UAM traffic scenarios. Based on Edie's definition, traffic flow theory is then applied with a near-stationary traffic condition filtering method to construct the FD. To account for real-world disturbances and modeling uncertainties, we further conduct physical experiments on a reduced-scale testbed using Bitcraze Crazyflie drones. Both simulation and physical experiment trajectory data are collected and organized into the UAMTra2Flow dataset, which is analyzed using the proposed framework. Preliminary results indicate that classical FD structures for ground transportation, especially the Underwood model, are applicable to UAM systems. Notably, FD curves obtained from physical experiments exhibit deviations from simulation-based results, highlighting the importance of experimental validation. Finally, results from the reduced-scale testbed are scaled to realistic operating conditions to provide practical insights for future UAM traffic systems. The dataset and code for this paper are publicly available at \url{https://github.com/CATS-Lab/UAM-FD}. (The source code will be released upon acceptance of this paper.)
}

\KEYWORDS{Urban Air Mobility; Fundamental Diagram; Traffic Flow}
\maketitle

\section{Introduction}

Recent advances in hardware, such as lightweight composite materials, high energy-density batteries, and precision navigation technologies, have significantly improved the performance and expanded the applications of unmanned aerial vehicles (UAVs) or drones across various fields \citep{otto2018optimization,frachtenberg2019practical,zhou2023exact}. Among these applications, Urban Air Mobility (UAM), which uses drones to transport passengers or cargo at low altitudes in urban areas, has attracted notable attention due to its benefit of reducing travel and delivery time \citep{li2024urban,ahmed2024state}. Although real-world deployment of UAM systems is still in its early stages, the growing number of drones operating in urban environments is expected to lead to congestion phenomena similar to those observed in ground transportation networks. However, the characteristics of UAM network congestion, which are an essential foundation for designing effective traffic management strategies, still remain poorly understood. Without a thorough analysis of UAM traffic properties, the operational efficiency advantages of UAM could diminish, and in some cases, even introduce potential safety risks.

In traffic flow theory, \textbf{fundamental diagram} (FD) characterizes the fundamental relationship between key traffic measures (e.g., density vs. flow) under certain driver behavior/operational rules in a roadway channel (e.g., \cite{greenshields1935study, greenberg1959analysis}) or in a network (e.g., \cite{daganzo2008analytical, saberi2014estimating, ambuhl2020functional}). They help understand surface road traffic evolution over time and space, and have been widely applied to various research problems and engineering applications \citep{qu2017stochastic,baer2019threshold,batista2019regional,chen2024iterative}. In the literature, various functional forms for not only traditional human-driven vehicles but also advanced transportation technologies have been adopted, such as multi-regime linear, polynomial, and exponential \citep{wu2002new,zheng2016modeling,shi2021constructing,amirgholy2017modeling,zhong2018boundary}. However, the majority of existing studies focus on one-dimensional (1D) highway settings, with only limited efforts devoted to higher-dimensional spatial environments. For example, several studies have examined pedestrian traffic in 2D walkway settings. \cite{Daamen2005pedestrain} developed traffic flow theory for pedestrians in 2D domains. \cite{flotterod2015bidirectional} derived a bidirectional FD from microscopic principles and defined direction-specific flow-density relationships with bounded dynamics. \cite{Serge2017MFDpedestrain} proposed a macroscopic FD for pedestrian networks. For a comprehensive review of pedestrian FDs, readers are referred to \cite{vanumu2017fundamental}. Building upon these prior studies, researchers have only recently begun to investigate the development of FDs for high-dimensional UAM systems. Recently, \cite{cummings2021emergence}, \cite{cummings2023measuring}, and \cite{cummings2024comparing} measured the density and flow data of drones in unstructured and structured 3D domains. \cite{battista2017modeling} considered the influence of different varieties of wind, such as tailwinds, headwinds, and winds of consistent strength for drone operations while studying the FD in UAM. \cite{haddad2021traffic} calibrated the FD for UAM using a different collision avoidance mechanism based on the model proposed by \cite{xue2019scenario}. However, as far as we know, all existing UAM FD results are based on only theoretical analyses or simulation-based approaches, and they have not yet been validated with physical experiments. Consequently, many practical factors are excluded from consideration, such as the uncertainties inherent in practical scenarios, the mechanics and dynamics of drones, and the effects of individual drone control algorithms.

\begin{figure}
    \centering
    \includegraphics[width=1\linewidth]{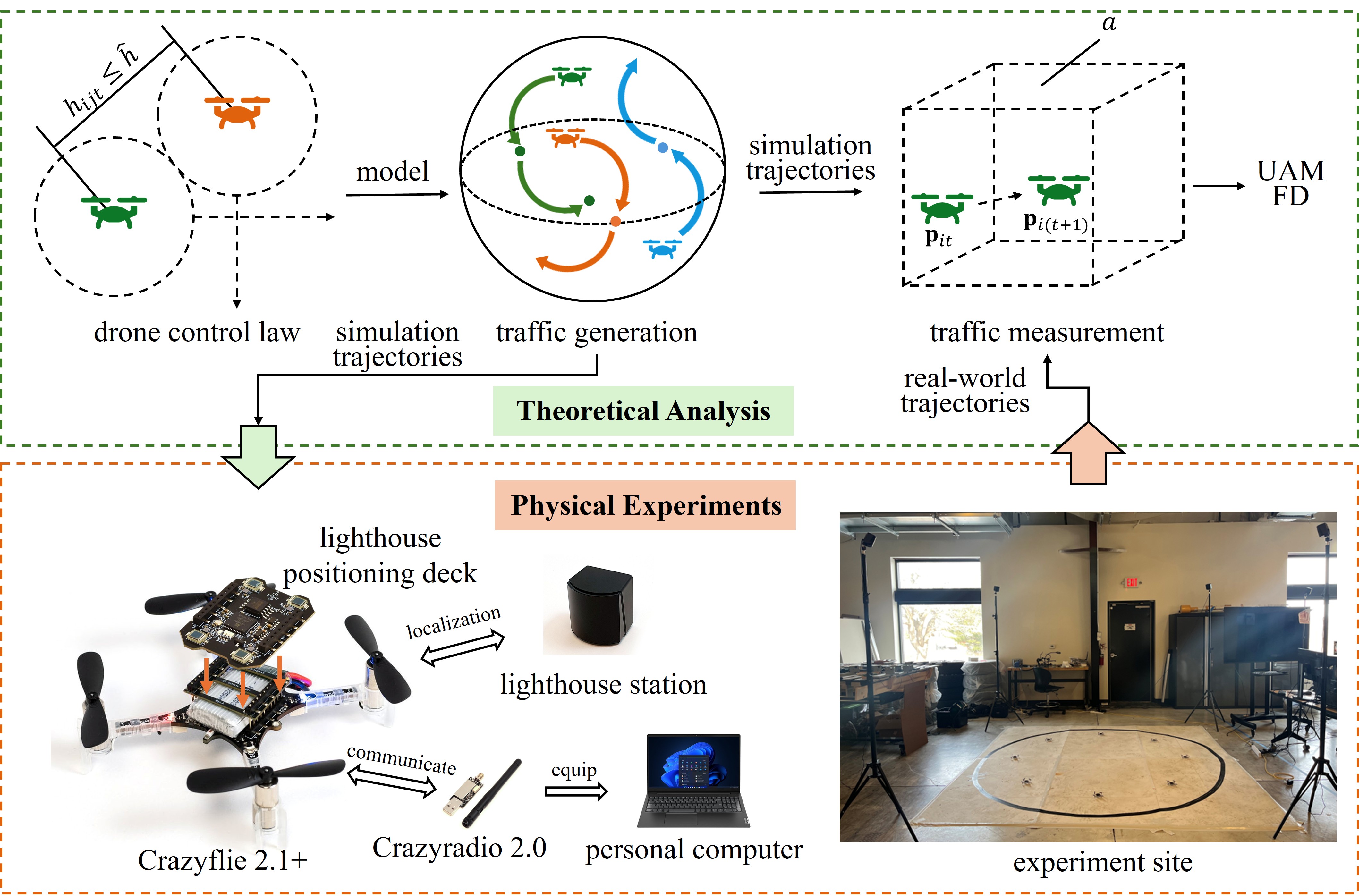}
    \caption{Overview of the proposed UAM FD construction method.}
    \label{fig:framework}
\end{figure}

To effectively support informed decision-making in UAM traffic, it is crucial to develop UAM FDs based on real-world drone traffic data. This paper proposes a general framework for constructing the FD for UAM, incorporating theoretical analysis and physical experiments. Figure~\ref{fig:framework} illustrates the overall framework of the proposed method. 
On the theoretical side, we first design two control laws inspired by the unsignalized intersection control, namely stop sign control and roundabout control. These control laws focus on collision avoidance when conflicts arise in the predicted future trajectories of drones, which are the main factors leading to congestion in the UAM. Based on these control laws, we further develop simulation-based traffic generation methods to create diverse traffic scenarios and improve the generalizability of our findings. Finally, using traffic flow theory, we identify key measurements such as density and flow for near-stationary traffic conditions, and calibrate the FD of UAM traffic, which captures the macroscopic relationships and supports the analysis of UAM traffic properties.
On the experimental side, to incorporate real-world disturbances such as drone dynamics and system delays, we conduct physical experiments using a reduced-scale platform equipped with Bitcraze Crazyflie drones. Trajectories generated from simulation are used as reference inputs and transmitted as control commands to the drone swarm. The drones' actual movements are then recorded by the positioning system, and the collected trajectory data are analyzed using FD theory to validate and refine the theoretical and simulation-based results.
Preliminary calibration results indicate that the traditional FD structure remains applicable for describing the traffic flow characteristics of UAM. Moreover, some conclusions commonly observed in ground transportation are also applicable to UAM, such as the impact of different control laws and parameter settings on variations in the FD curve, and the significant enhancement of traffic capacity with smaller safety spacing settings. Besides, it was observed that the FD curves derived from physical experiments tend to shift rightward compared to those obtained from simulations, which highlights the need for physical experiments to corroborate and improve findings from pure theoretical or simulation studies. 
Finally, to obtain further insights into real-world UAM traffic, we scale the results from the reduced-scale platform to realistic operating conditions, which can serve as a benchmark reference for future studies and applications.

To the best of our knowledge, this study is the first to construct FDs for UAM traffic based on physical experiment data. Our work contributes to the literature in the following aspects:
\begin{itemize}
   \item This paper proposes \textbf{a general framework for constructing the FD of UAM traffic}. The theoretical framework integrates the design of two drone control laws, a simulation-based UAM traffic generation method, and tailored traffic flow measurement and near-stationary traffic condition filtering methods for UAM systems. The proposed framework is modular and can be easily adapted to different UAM application scenarios.
   
   \item A reduced-scale UAM testbed is developed, and \textbf{a real-world UAM trajectory traffic flow dataset, UAMTra2Flow}, is collected to support UAM traffic studies. As the first publicly available dataset derived from physical experiments on UAM traffic, it encourages the research community to consider the impact of real-world disturbances and helps generalize existing findings to more practical operating conditions.
   
   \item Using the proposed framework and dataset, we \textbf{construct UAM traffic FDs based on physical experiment data}. Comparative experiments are conducted to analyze the results of different FD models and factors that influence the FD, as well as the differences between simulation-based and physical-experiment-based FDs. In addition, results from the reduced-scale testbed are scaled to realistic operating conditions to provide more insights and serve as a benchmark reference.
\end{itemize}

The remainder of the paper is as follows. Section \ref{sec:method} presents the theoretical analysis method for traffic generation, traffic measurement, and drone control law design. Section \ref{sec:collect} describes the details of physical experiments and shows the datasets collected by this study. Section \ref{sec:experiments} discussed the experimental results. Section \ref{sec:conclusion} concludes this paper and suggests future research directions.

\section{Theoretical framework}
\label{sec:method}

\begin{figure}
    \centering
    \includegraphics[width=1\linewidth]{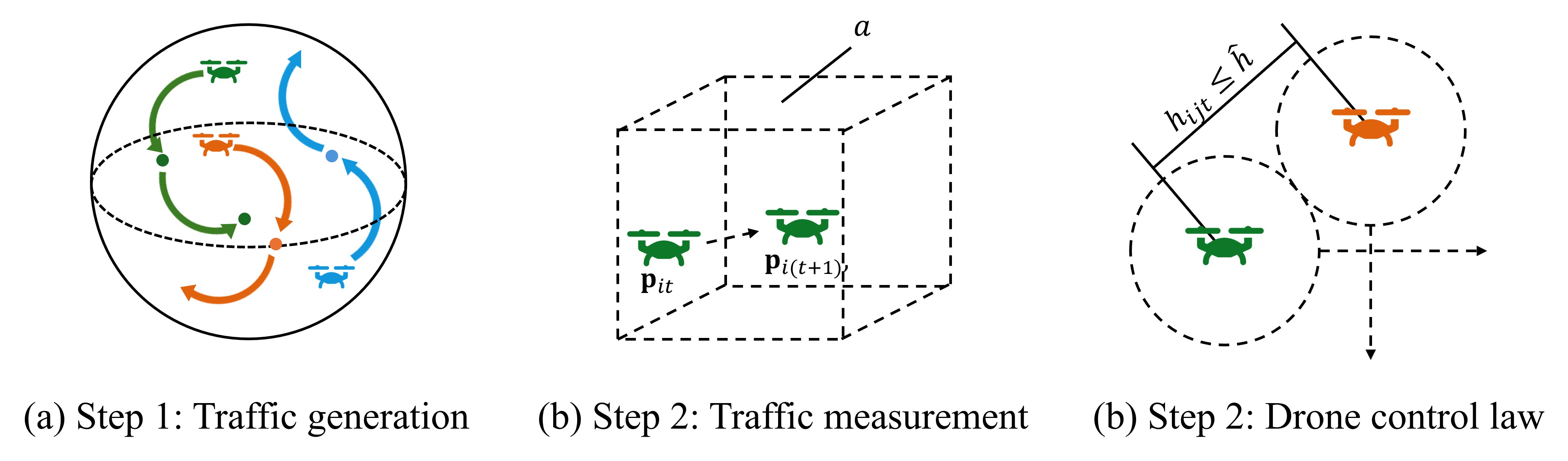}
    \caption{Three steps of the theoretical framework of the UAM FD construction based on simulation.}
    \label{fig:method}
\end{figure}

This section presents a theoretical framework for UAM traffic data collection based on simulation, as well as the construction of FDs for UAM traffic. To introduce the key concepts and notations clearly, we follow the structure shown in Figure~\ref{fig:method}, which outlines three main steps: traffic generation, traffic measurement, and drone control law design. In Step 1, we define the experimental setup and generate drone trajectories under diverse traffic scenarios to cover a broad range of traffic conditions. Step 2 applies the traditional traffic flow theory for ground transportation to the UAM, especially how to measure the flow and density, identify near-stationary traffic conditions, and calibrate the FDs. In Step 3, we discuss the control strategies for individual drones, with an emphasis on collision avoidance algorithms when multiple drones' trajectories have conflict points. The trajectory scenarios generated through these steps are then used as input for the drone control law in the physical experiments, as detailed in Section~\ref{sec:collect}. It is important to note that this study demonstrates a benchmark implementation of the three-step framework. Future work may explore extended variations within each step to design and conduct more comprehensive experiments.

\subsection{General traffic information and traffic generation}
\label{sec:definition}

Considering a given set of $I$ drones, denoted as $\mathcal{I}=\{1,2,...,I\}$, each drone performs a continuous sequence of flights that simulate real-world tasks, including package delivery, infrastructure inspection, and passenger transport. Each flight has a defined origin and destination (OD). To simplify the analysis and mitigate the boundary effects, drone movement is modeled on the surface of a spherical 3D space, denoted as $\mathcal{A}$. As shown in Figure~\ref{fig:framework}(a), we assume that all OD points are located on the surface, which implies that all trips are performed at the same altitude. 
This assumption is consistent with both prior literature and practical implementations, where urban airspace is commonly divided into multiple layers at different altitudes and drones typically operate within one of the designated layers. For example, recent airspace management reforms in China propose dividing urban low-altitude airspace into layers with 15-meter vertical intervals, allowing drones to operate within separate horizontal planes \citep{cctv2024uam}. A similar urban airspace structure based on vertical layers is also discussed in \cite{haddad2021traffic}.

In this UAM system, each drone starts from one origin and visits multiple consecutive destinations. During the flight, all drones keep the same desired cruising speed $\bar{v}$. The trajectory for each drone can be represented as a series of consecutive position data points in consecutive time steps $\mathcal{T}=\{0,1,...,T\}$ with equal interval $\Delta t$, where $T$ is the total number of time steps. As shown in Figure~\ref{fig:framework}(b), the position vector for drone $i \in \mathcal{I}$ at time $t\in \mathcal{T}$ is denoted as $\mathbf{p}_{it}\in\mathbb{R}^3$. Denote the OD pairs for drone $i$ as $(\mathbf{p}^{\text{O}}_{in}, \mathbf{p}^{\text{D}}_{in})$, where $n \in \mathcal{N} := \{1,\ldots,N\}$ indexes the sequence of flights and $N$ is the total number of flights. Consecutive flights are connected, so that the destination of flight $n$ coincides with the origin of flight $n+1$, i.e., $\mathbf{p}^{\text{D}}_{in} = \mathbf{p}^{\text{O}}_{i(n+1)}$ for all $n \in \{1,\ldots,N-1\}$.
The flight trajectory between any OD pair is assumed to follow the great-circle path on the spherical airspace. Given an origin point $\mathbf{p}^{\text{O}}_{in}$ and a destination point $\mathbf{p}^{\text{D}}_{in}$, we generate intermediate positions along the unique great-circle arc connecting them using spherical linear interpolation (slerp) \citep{slerp}. The slerp operator returns the point corresponding to a normalized arc-length parameter $\ell\in[0,1]$ between two locations on the sphere. Therefore, we first compute the great-circle distance $L$ between $\mathbf{p}^{\text{O}}_{in}$ and $\mathbf{p}^{\text{D}}_{in}$ using the spherical law of cosines \citep{spherical_law_of_cosines}. Since the drone advances a physical distance of $\bar{v}\,\Delta t$ at each simulation step, the interpolation parameters then follow $\ell_h = h\,\bar{v}\,\Delta t / L,\; h\in\ \mathcal{H} := \{0,\ldots,H\}$, where $H=\lceil L/(\bar{v}\,\Delta t)\rceil$. Applying slerp to each $\ell_i$ yields a time-consistent sequence of waypoints that discretizes the great-circle trajectory.

To represent a wide range of UAM traffic conditions, we consider different demand distributions, representing the geographic spread of customers' ODs. Three demand distribution scenarios are included: (1) randomly generated OD pairs continuously distributed on the spherical surface, mimicking online delivery with distributed customers; (2) origins and destinations are separated in different zones, respectively, mimicking segregated demand and destination patterns such as daily commute between residential and workplace areas; and (3) Random OD pair selection among fixed stations, where the eight stations are placed at the vertices of a cube inscribed in the spherical airspace, mimicking station-based delivery or station-to-station travel services.

\subsection{Measuring UAM traffic}

This section introduces the methods used to measure UAM traffic conditions based on traffic flow theory. Specifically, we first introduce the definitions of flow and density for UAM traffic, adapted from Edie’s generalized framework \citep{edie1963discussion}. However, the resulting flow--density data may include non-stationary traffic conditions, which can lead to biased or misleading FDs \citep{cassidy1998bivariate,blandin2013phase,bramich2022fitting}. 
To extract flow--density data corresponding to near-stationary traffic conditions, we develop a region partition method, and apply a time window selection procedure and a data filtering method adapted from \citet{cassidy1998bivariate}. Finally, we briefly introduce the FD calibration procedure.

\subsubsection{Measurements definition.}\label{sec:measure}

Traditionally, the fundamental characteristics of ground vehicle traffic flow include density, flow rate, and speed. In 1D space, these metrics can be directly measured from trajectory data using Edie’s generalized definitions \citep{edie1963discussion}. However, when the spatial dimension increases to 2D or even 3D, the measurement approach requires modification. Prior studies, such as \cite{saberi2014exploring} and \cite{cummings2021emergence}, have proposed extensions of Edie’s definitions to higher-dimensional traffic systems. This study adopts their approach.

Specifically, we consider a predefined 2D measurement region $a \subseteq \mathcal{A}$ and a time window $w \subseteq \mathcal{T}$. The volume of this time-space region is calculated as $|a| \cdot |w| \Delta t $, where $|a|$ is the area of region $a$ and $|w|$ is the number of time steps of the time window $w$. For each drone $i \in \mathcal{I}$ that enters region $a$ during $w$, its travel distance and time spent within the region are denoted as $s_{iaw}$ and $t_{iaw}$, respectively. Then, the density and flow of drones in $a$ during time window $w$, denoted by $k_{aw}$ and $q_{aw}$ respectively, can be calculated as:
\begin{align}
    k_{aw} =\frac{\sum_{i \in \mathcal{I}} t_{iaw}}{|a| \cdot |w| \Delta t} \label{eq:density}\\
    q_{aw} =\frac{\sum_{i \in \mathcal{I}} s_{iaw}}{|a| \cdot |w| \Delta t}. \label{eq:flow}
\end{align}

\begin{figure}
    \centering
    \includegraphics[width=0.5\linewidth]{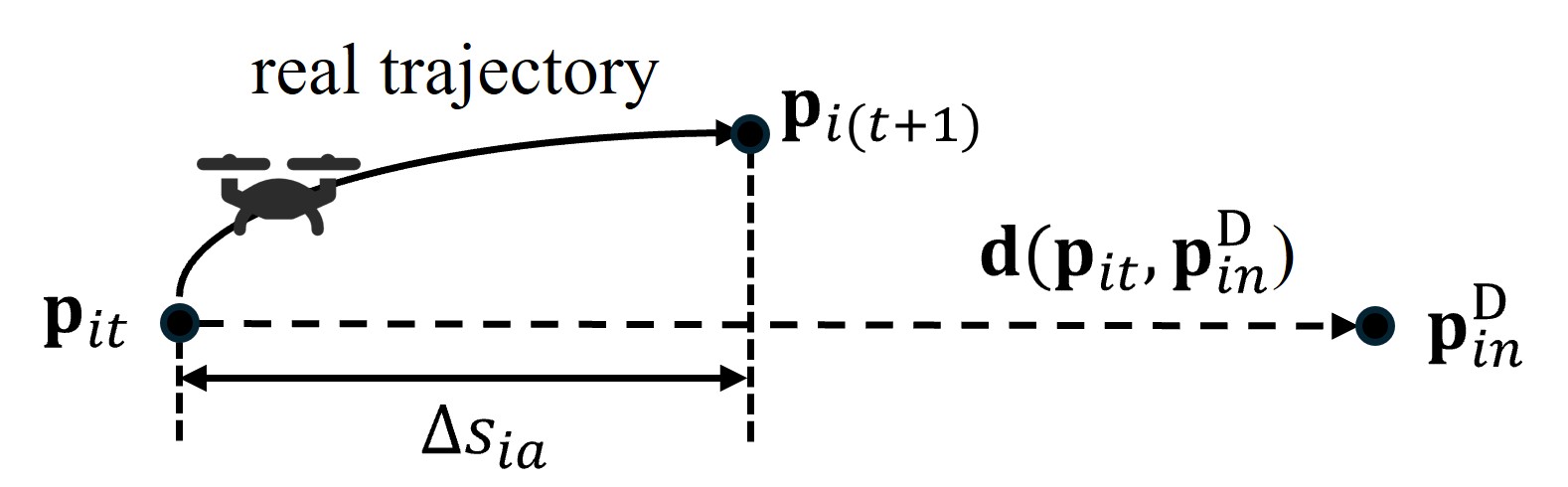}
    \caption{Illustration of the effective distance.}
    \label{fig:dis}
\end{figure}

However, unlike the traditional car-following behavior observed on 1D highways, drones in this scenario may not only change their speed but also deviate from their planned trajectories to avoid potential collisions with other drones approaching from various directions. These potential detours must be considered in the analysis. Therefore, the measured distance $s$ is transferred from the travel distance to the effective distance, i.e., the distance traveled by drone $i$ projected onto the direction toward its next destination. Figure~\ref{fig:dis} illustrates this definition. The dashed line indicates the nominal direction from the current position $\mathbf{p}_{it}$ to the destination $\mathbf{p}^{\text{D}}_{in}$, while the solid line shows the actual trajectory from $\mathbf{p}_{it}$ to $\mathbf{p}_{i(t+1)}$. Only the projection of the actual trajectory onto the nominal direction contributes to the effective distance $s_{iaw}$. 

\subsubsection{Region partition.} \label{sec:partition}

As discussed in the literature, FDs calibrated from empirical data often differ from their theoretical counterparts, and one important reason is the presence of non-stationary traffic conditions \citep{cassidy1998bivariate,blandin2013phase,bramich2022fitting}. For example, if the entire spherical airspace $\mathcal{A}$ is treated as a single measurement region, the resulting flow and density data may contain mixed traffic conditions, with some drones operating under free-flow conditions while others experience congestion. As a result, the corresponding flow--density pairs do not represent a single point on the FD, but rather a linear combination of multiple traffic states on the FD. To address this issue, we apply a set of methods to identify near-stationary traffic conditions. These methods are introduced in this and the following sections.

The first method we applied is region partition. This method aims to compute flow–density pairs not from a single global region but from a set of smaller and homogeneous subregions, which reduces the likelihood that fundamentally different traffic states are aggregated within the same measurement region. Specifically, the spherical airspace $\mathcal{A}$ is partitioned into $M$ non-overlapping regions $a_1, a_2, \ldots, a_M \subset \mathcal{A},$ such that $|a_n| = |a_m|$, $a_n \cap a_m = \emptyset, \forall n, m \in \mathcal{M} := \{1,...,M\}, n \neq m$, and $\bigcup_{i=1}^{M} a_i = \mathcal{A}$. To construct these regions, we adopt an equal-angle discretization of the spherical surface, parameterized by an integer $\bar{m}$. The sphere is divided into $\bar{m}$ segments along the polar angle $\theta\in[0,\pi]$ and $\bar{m}$ segments along the azimuth angle $\varphi\in[0,2\pi]$, resulting in $M=\bar{m}^2$ disjoint subregions. For each drone position $\mathbf{p}$, its spherical coordinates $(\theta,\varphi)$ are computed, and the corresponding subregion index is obtained by uniformly quantizing both angular dimensions into $m$ intervals.


\subsubsection{Time window selection.}
\label{sec:time}

The next step is to select appropriate time windows $w$. To identify near-stationary traffic conditions, we adapt the time window selection approach proposed by \citet{cassidy1998bivariate}, originally developed for 1D highway traffic, to the UAM setting. The objective is to identify time windows during which the arrival process into a spatial subregion satisfies a quasi-linear arrival pattern.

Consider a spatial subregion $a_m$, $m \in \mathcal{M}$. An entry event into $a_m$ is defined to occur when a drone moves from outside the subregion to inside between two consecutive discrete time steps. Based on these entry events, the cumulative number of arrivals into subregion $a_m$ up to time step $t$ is given by
\begin{align}
    E_{mt} =
    \sum_{i \in \mathcal{I}}
    \sum_{t' = 1}^{t}
    \mathbb{I}\!\left(
    \mathbf{p}_{i(t'-1)} \notin a_m
    \;\land\;
    \mathbf{p}_{it'} \in a_m
    \right), \label{eq:nmt}
\end{align}
where $\mathbb{I}(\cdot)$ denotes the indicator function, which equals 1 if the condition is satisfied and 0 otherwise. Under quasi-stationary arrival conditions, $E_{mt}$ is expected to increase approximately linearly with time. To identify such conditions from discretized trajectory data, we examine arrival stability within sliding time windows. Consider a time window $w$ with fixed length. The discrete arrival rate at time $t \in w$ is defined as
\begin{align}
    f_{mt} = \frac{E_{mt} - E_{m(t-1)}}{\Delta t}. \label{eq:fmt}
\end{align}
A window is classified as exhibiting quasi-stationary arrivals if the arrival rate $f_{mt}$ shows no systematic temporal trend and only limited variability within $w$. Accordingly, for each $m \in \mathcal{M}$ and each time window $w \subset \mathcal{T}$, we compute the variance of $\{ f_{mt} \}_{t \in w}$. If the variance $\mathrm{Var}(f_{mt})$ exceeds a predefined threshold $\xi_f$, the corresponding window is excluded from further analysis.

\subsubsection{Non-stationary traffic conditions filtering.}
\label{sec:filter}

Stable arrival processes alone do not guarantee near-stationary traffic conditions, as traffic states within a subregion may continue to evolve due to changes in speed, spacing, or local interactions. Therefore, a filtering procedure adapted from \citet{cassidy1998bivariate} is applied to further identify time windows in which the traffic state within each subregion is close to local equilibrium.

For each subregion $a_m$ and each candidate time window $w$ obtained from Section~\ref{sec:time}, we jointly examine cumulative arrivals and cumulative occupancy. Denote $w_t$ as the time window $[0,t]$, the cumulative occupancy is then defined as
\begin{align}
    C_{mt} = \sum_{i\in \mathcal{I}} t_{iaw_t} \label{eq:cmt}
\end{align}
which represents the total time spent by all drones inside subregion $a_m$ up to time step $t$. Within a time window $w=[t_s, t_e]$, both cumulative sequences are reset at the interval start by:
\begin{align}
    \tilde E_{mt} &= E_{mt} - E_{mt_s}, \\
    \tilde C_{mt} &= C_{mt} - C_{mt_s}.
\end{align}
To enable direct comparison, the cumulative occupancy sequence is rescaled so that the two sequences have the same magnitude at the interval end:
\begin{align}
    \bar C_{mt} &= \frac{\tilde E_{mt_e}}{\tilde C_{mt_e}} \, \tilde C_{mt}.
\end{align}

A common linear trend is removed from both cumulative sequences to isolate short-term fluctuations. Let $b_0$ denote the linear trend coefficient over the time window $[t_s, t_e]$, estimated from the cumulative arrivals as
\begin{align}
    b_0 = \frac{\tilde N_{mt_e} - \tilde N_{mt_s}}{(t_e - t_s)\Delta t}.
\end{align}
The detrended sequences are then defined as
\begin{align}
    W^{\mathrm{E}}_{t} &= \tilde E_{mt} - b_0 (t - t_s)\Delta t, \\
    W^{\mathrm{C}}_{t} &= \bar C_{mt} - b_0 (t - t_s)\Delta t. \label{eq:detrend}
\end{align}

Under near-stationary traffic conditions, cumulative arrivals and cumulative occupancy are expected to remain approximately proportional, implying that their detrended fluctuations exhibit similar temporal patterns. Accordingly, a candidate window $w$ is classified as near-stationary if the detrended sequences $W^{\mathrm{E}}_{t}$ and $W^{\mathrm{C}}_{t}$ show strong consistency within the window, quantified by the correlation coefficient $\rho_{mw}$ between the two detrended sequences within the window. If the correlation is lower than a predefined threshold $\xi_{\rho}$, the data within this window are excluded from FD calibration.

\subsubsection{FD calibration.}

With the above definition, we can obtain a set of flow--density pairs $(q_{aw}, k_{aw})$. Based on these data, FD curves are developed via regression using several well-known functional forms from the traffic flow literature. Specifically, five classical models are considered, including the Greenshields model \citep{greenshields1935study}, the Greenberg model \citep{greenberg1959analysis}, the Underwood model \citep{underwood1960speed}, and the Drake model \citep{drake1967statistical}. All models are calibrated using the least-squares method (LSM), which estimates model parameters by minimizing the sum of squared residuals. The fitting performance of different models is compared and discussed in Section~\ref{sec:experiments}.



\subsection{Drone control laws}

In traffic flow theory, the control law defines the microscopic behavior of individual vehicles as they interact with one another and respond to environmental constraints. The macroscopic properties characterized by the FD are therefore intrinsically determined by these microscopic rules. In the context of UAM, drone behavior in congested environments, such as deceleration, yielding, or rerouting, is primarily dictated by collision-avoidance strategies \citep{tang2021automated,tang2023predeparture}. Accordingly, the control laws examined in this study explicitly focus on collision-avoidance mechanisms.

Although the robotics community has developed a wide range of collision-avoidance algorithms for UAV \citep{rezaee2024comprehensive}, these methods are largely designed for emergent, local obstacle-avoidance scenarios and focus on guaranteeing safety for individual drones. In contrast, almost no studies have investigated avoidance strategies from a traffic-control perspective, where collective behavior and macroscopic performance are of primary interest. Because this study aims to examine the macroscopic properties of UAM traffic, we consider two control laws inspired by the unsignalized intersection control: a stop-and-yield strategy and a circular-detour strategy, following the ideas of the stop-sign and the roundabout operations, respectively. As shown in Figure~\ref{fig:framework}(c), the spacing between drone $i$ and drone $j\in\mathcal{I}$ at time $t$ is denoted by $h_{ijt}=\|\mathbf{p}_{it}-\mathbf{p}_{jt}\|$. For both control laws, a predefined safe spacing $\hat{h}$ is used to identify potential conflicts. The strategy is triggered between drones $i$ and $j$ at time $t\in\mathcal{T}$ whenever $h_{ijt}\le \hat{h}$. Then, we present the details of the two control laws.

\begin{figure}
    \centering
    \includegraphics[width=1.\linewidth]{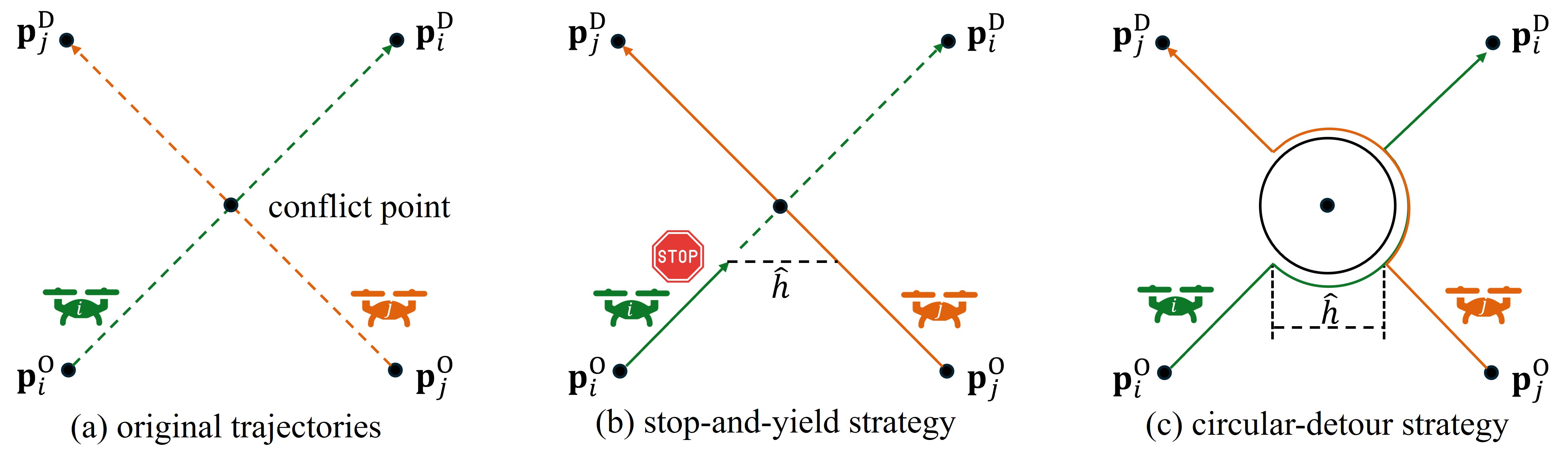}
    \caption{An example of the stop-and-yield and circular-detour strategies.}
    \label{fig:control}
\end{figure}

The first control law is the \emph{stop-and-yield strategy}, which mimics the stop-sign. As shown in the example for two drones in Figure~\ref{fig:control}(a) and (b), once the strategy is triggered, one of the two drones halts and waits until the spacing exceeds $\hat{h}$. Unlike ground transportation systems, UAM traffic lacks fixed intersection locations, making it impossible to determine right-of-way based on arrival order. To avoid situations in which multiple drones wait for one another, a priority rule is introduced: the lower-priority drone yields to the higher-priority one. In Figure~\ref{fig:control}(b), the left (green) drone $i$ has lower priority, and therefore, after following the solid trajectory segment, it yields to drone $j$ and waits until drone $j$ passes, before continuing along the dashed trajectory toward its destination. A similar yield principle is applied in scenarios with more than two drones. For simplicity, priorities in this study are assigned using drone indices. In practical implementations, however, priorities could be determined dynamically based on drone states such as positions, velocities, and OD information, or could be randomized to emulate decentralized conflict resolution without predefined rules.

The second control law is the \emph{circular-detour strategy}, inspired by the operation of roundabouts in roadway systems. When a potential collision is detected in the local neighborhood, a virtual circular roundabout is implicitly formed, and the drones resolve the conflict by rotating their intended directions while maintaining continuous flight. As shown in Figure~\ref{fig:control}(c), drones $i$ and $j$ both enter a virtual roundabout constructed between them and travel along circular detour trajectories to avoid collision. Specifically, at time $t$, if drone $i$ is navigating toward its destination $\mathbf{p}^{\text{D}}_{in}$, its nominal direction of motion is given by the unit vector $\mathbf{d}(\mathbf{p}_{it}, \mathbf{p}^{\text{D}}_{in})$. To explore alternative rotation directions, this nominal direction is rotated around the outward normal $\hat{\mathbf{p}}_{it}=\mathbf{p}_{it}/\|\mathbf{p}_{it}\|$ using Rodrigues rotations \citep{pina2011rotations}. The rotation follows the right-hand rule with respect to $\hat{\mathbf{p}}_{it}$, ensuring a consistent orientation on the local tangent plane. A set of candidate rotation angles $\Psi$ is generated by uniformly sampling the interval $[0,2\pi]$.
For each sampled angle $\psi \in \Psi$, the rotated direction is obtained as
\begin{align}
    \mathbf{d}_{\text{rot}}(\psi)
    = \mathbf{d}(\mathbf{p}_{it}, \mathbf{p}^{\text{D}}_{in})\cos\psi
    + (\hat{\mathbf{p}}_{it}\times\mathbf{d}(\mathbf{p}_{it}, \mathbf{p}^{\text{D}}_{in}))\sin\psi
    + \hat{\mathbf{p}}_{it}\bigl(\hat{\mathbf{p}}_{it}\cdot\mathbf{d}(\mathbf{p}_{it}, \mathbf{p}^{\text{D}}_{in})\bigr)(1-\cos\psi),
    \label{eq:rodrigues}
\end{align}
Instead of enforcing a hard feasibility constraint for all neighboring drones, the strategy evaluates the preference induced by each neighbor individually. For each neighboring drone $j$, a tangential steering preference is constructed based on the relative direction $\mathbf{d}(\mathbf{p}_{it},\mathbf{p}_{jt})$, encouraging lateral deviation while maintaining separation. The candidate direction that best aligns with each neighbor’s tangential preference is first identified. These neighbor-wise preferred directions are then aggregated using inverse-distance weights, yielding a consensus steering direction. Finally, the rotated displacement $\mathbf{d}_{\text{rot}}(\psi)$ that is most aligned with this aggregated preference is selected for execution. Although this strategy does not explicitly optimize path length, it provides a robust way when multiple neighboring conflicts occur simultaneously.


\subsection{Algorithm overview}
\label{sec:overview}

With all the components described in the previous sections, this subsection summarizes the overall procedures for UAM traffic simulation and FD calibration. Algorithm~\ref{alg:framework} outlines the UAM traffic simulation pipeline, which iteratively updates drone states according to OD assignments and the specified control laws, thereby generating full-trajectory datasets. Using these simulated trajectories, we can directly obtain the corresponding FD of the simulated traffic flow.
Algorithm~\ref{alg:fd} then converts the trajectory data into macroscopic flow–density measurements by computing Edie-based density and flow over the spherical subregions and subsequently fitting the FD. In addition to simulation-based FD construction, the same procedure in Algorithm~\ref{alg:fd} can be applied to the real-world flight trajectories collected through the physical experiments presented in the next section, yielding the empirical FD of real UAM traffic.

\begin{algorithm}[t]
\caption{UAM Traffic Simulation Framework}
\label{alg:framework}
\begin{algorithmic}[1]
    \STATE \textbf{Input:} number of drones $I$, time step $\Delta t$, maximum steps $T$, speed $\bar{v}$, safe spacing $\hat{h}$, scenario type, control law type.
    \FORALL{$i \in \mathcal{I}$}
        \STATE Generate OD pairs $\{(\mathbf{p}^{\text{O}}_{in},\mathbf{p}^{\text{D}}_{in})\}_{n\in \mathcal{N}}$ according to the scenario type.
        \STATE Initialize drone position $\mathbf{p}_{i0}$.
    \ENDFOR
    \FORALL{$t \in \mathcal{T}$}
        \FORALL{$i \in \mathcal{I}$}
            \STATE Log each drone's current position $\mathbf{p}_{it}$ and destination $\mathbf{p}^{\text{D}}_{in}$.
            \STATE Compute distances $h_{ijt}=\|\mathbf{p}_{it}-\mathbf{p}_{jt}\|$ to neighboring drones.
            \IF{$\min_j h_{ijt} \le \hat{h}$}
                \IF{control law = stop-and-yield}
                    \STATE Drone $i$ stops at $\mathbf{p}_{it}$.
                \ELSIF{control law = circular-detour}
                    \STATE Generate candidate rotation angles $\Psi$ by uniformly sampling $[0,2\pi]$.
                    \STATE For each $\psi \in \Psi$, compute $\mathbf{d}_{\text{rot}}(\psi)$ using Eq.~\eqref{eq:rodrigues}.
                    \STATE For each neighboring drone $j$, identify the candidate angle that best aligns with its tangential preference.
                    \STATE Aggregate the neighbor-wise preferred directions using inverse-distance weights.
                    \STATE Select the rotated direction that is most aligned with the aggregated preference.
                    \STATE Move drone $i$ along the selected $\mathbf{d}_{\text{rot}}(\psi)$.
                \ENDIF
            \ELSE
                \STATE Move toward $\mathbf{p}^{\text{D}}_{in}$ via nominal slerp step.
            \ENDIF
            \STATE Terminate if drone $i$ reach final destination $\mathbf{p}^{\text{D}}_{iN}$.
        \ENDFOR
    \ENDFOR
    \STATE \textbf{Output:} drone trajectories $\{\mathbf{p}_{it}\}_{i \in \mathcal{I},t\in \mathcal{T}}$ and OD pairs $\{(\mathbf{p}^{\text{O}}_{in},\mathbf{p}^{\text{D}}_{in})\}_{i \in \mathcal{I},n\in \mathcal{N}}$.
    \end{algorithmic}
\end{algorithm}

\begin{algorithm}[t]
    \caption{FD Construction from Simulation or Experimental Data}
    \label{alg:fd}
    \begin{algorithmic}[1]
    \STATE \textbf{Input:} trajectories $\{\mathbf{p}_{it}\}_{i \in \mathcal{I},t\in \mathcal{T}}$, spherical partition parameter $\bar{m}$, window length $|w|$, thresholds $\theta_f$, $\theta_{\rho}$.
    \STATE Partition the spherical airspace $\mathcal{A}$ into identical subregions $\{a_m\}_{m\in\mathcal{M}}$ using $\bar{m}$.
    \FORALL{$m \in \mathcal{M}$}
        \STATE Compute cumulative arrivals $\{E_{mt}\}_{t\in \mathcal{T}}$ with Eq. \eqref{eq:nmt}.
        \STATE Slide time windows with fixed length $|w|$ to generate the candidate set $\mathcal{W}$.
        \FORALL{$w \in \mathcal{W}$}
            \STATE Compute arrival rates $\{f_{mt}\}_{t\in w}$ with Eq. \eqref{eq:fmt} and discard $w$ from $\mathcal{M}$ if $\mathrm{Var}(f_{mt}) > \xi_f$.
            \STATE Compute cumulative occupancy $\{C_{mt}\}_{t\in w}$ with Eq. \eqref{eq:cmt}.
            \STATE Detrend arrivals and occupancy to obtain $\{W^{\mathrm{E}}_{t}\}_{t\in w}$ and $\{W^{\mathrm{C}}_{t}\}_{t\in w}$ with Eq. \eqref{eq:detrend}.
            \STATE Compute correlation $\rho_{mw}$ between $\{W^{\mathrm{E}}_{t}\}_{t\in w}$ and $\{W^{\mathrm{C}}_{t}\}_{t\in w}$ and discard $w$ from $\mathcal{M}$ if $\rho_{mw} < \xi_{\rho}$.
        \ENDFOR
    \ENDFOR
    \STATE Compute flow--density pairs $\{(k_{a_mw}, q_{a_mw})\}_{m\in \mathcal{M},w\in \mathcal{W}}$ by Eqs.~\eqref{eq:density}--\eqref{eq:flow}.
    \STATE Fit FD model parameters via the LSM.
    \STATE \textbf{Output:} FD model parameters.
    \end{algorithmic}
\end{algorithm}

\section{Experiment design for data collection}
\label{sec:collect}

This section introduces the reduced-scale UAM testbed and the collected UAMTra2Flow dataset.

\subsection{Reduced-scale UAM testbed}
\label{sec:phy}

Our physical experiments were conducted using a reduced-scale experimental platform based on the Bitcraze Crazyflie 2.1+. Crazyflie 2.1+ is a versatile development platform for aerial robotics research and has been widely used in laboratory-based drone studies, as reported in the literature \citep{preiss2017crazyswarm,duisterhof2021tiny,hegde2024hyperppo}.

\subsubsection{Hardware Setup.}
Our hardware setup includes a personal computer equipped with a 3.2~GHz AMD Ryzen~7 7735HS CPU with Radeon Graphics and 16~GB of RAM, running the Ubuntu~22.04 operating system. Equipped with a Crazyradio~2.0, the computer is able to establish wireless communication with each Crazyflie, enabling real-time command transmission and data collection.

\begin{figure}
    \centering
    \includegraphics[width=1.\linewidth]{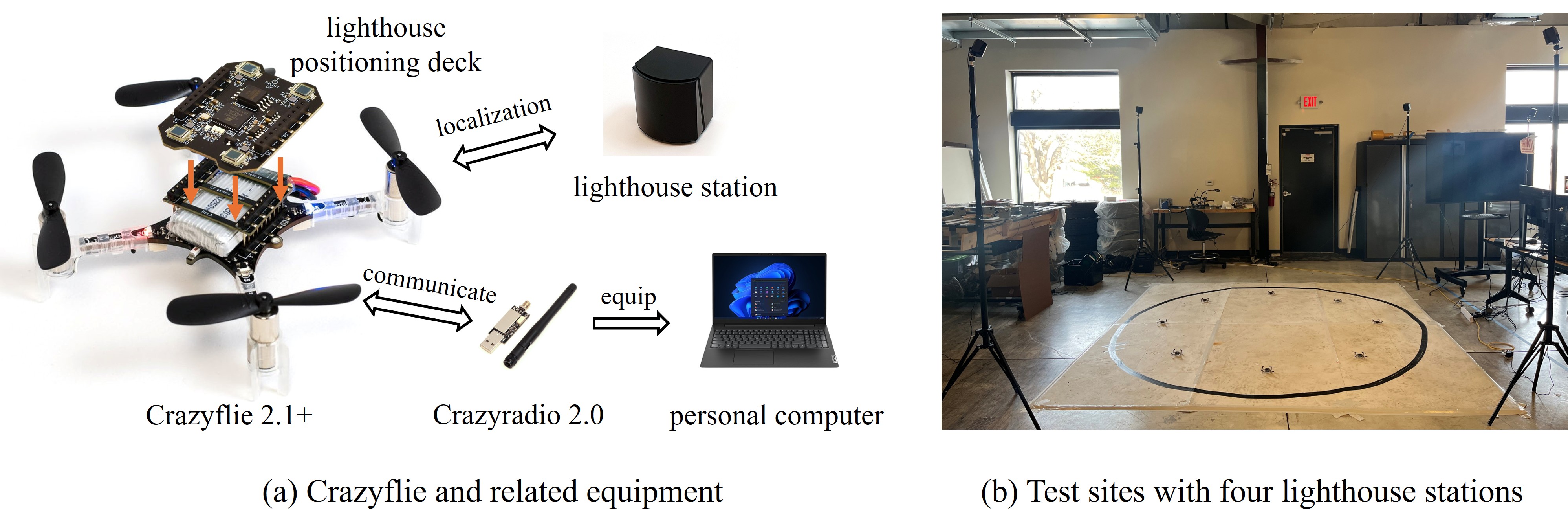}
    \caption{Hardware setup for the reduced-scale testbed and the deployment of the indoor test site.}
    \label{fig:resource}
\end{figure}

To obtain accurate position estimates, we use the Bitcraze-recommended indoor localization system based on lighthouse stations. The drones operate on a spherical surface of radius 1\,m. To ensure that the lighthouse system can reliably capture drone positions within this flight envelope, four lighthouse stations are installed at the four corners of a 3\,m~$\times$~3\,m rectangular area, each mounted at a height of approximately 2.5\,m. Each Crazyflie is equipped with a positioning deck that receives signals from the lighthouse stations and computes its 3D position. This setup enables robust and precise localization throughout the test space. The hardware configuration is summarized in Figure~\ref{fig:resource}.

\subsubsection{Software Architecture.} To achieve accurate control of the drone swarm, we employ the ROS~2-based Python package \texttt{Crazyswarm2} \citep{crazyswarm2}. Developed by the Intelligent Multi-Robot Coordination Lab, Crazyswarm2 is an open-source software framework specifically designed for controlling swarms of Bitcraze Crazyflie drones. It offers a comprehensive set of APIs that allow users to implement multi-drone control using Python scripts.

\begin{figure}
    \centering
    \includegraphics[width=0.8\linewidth]{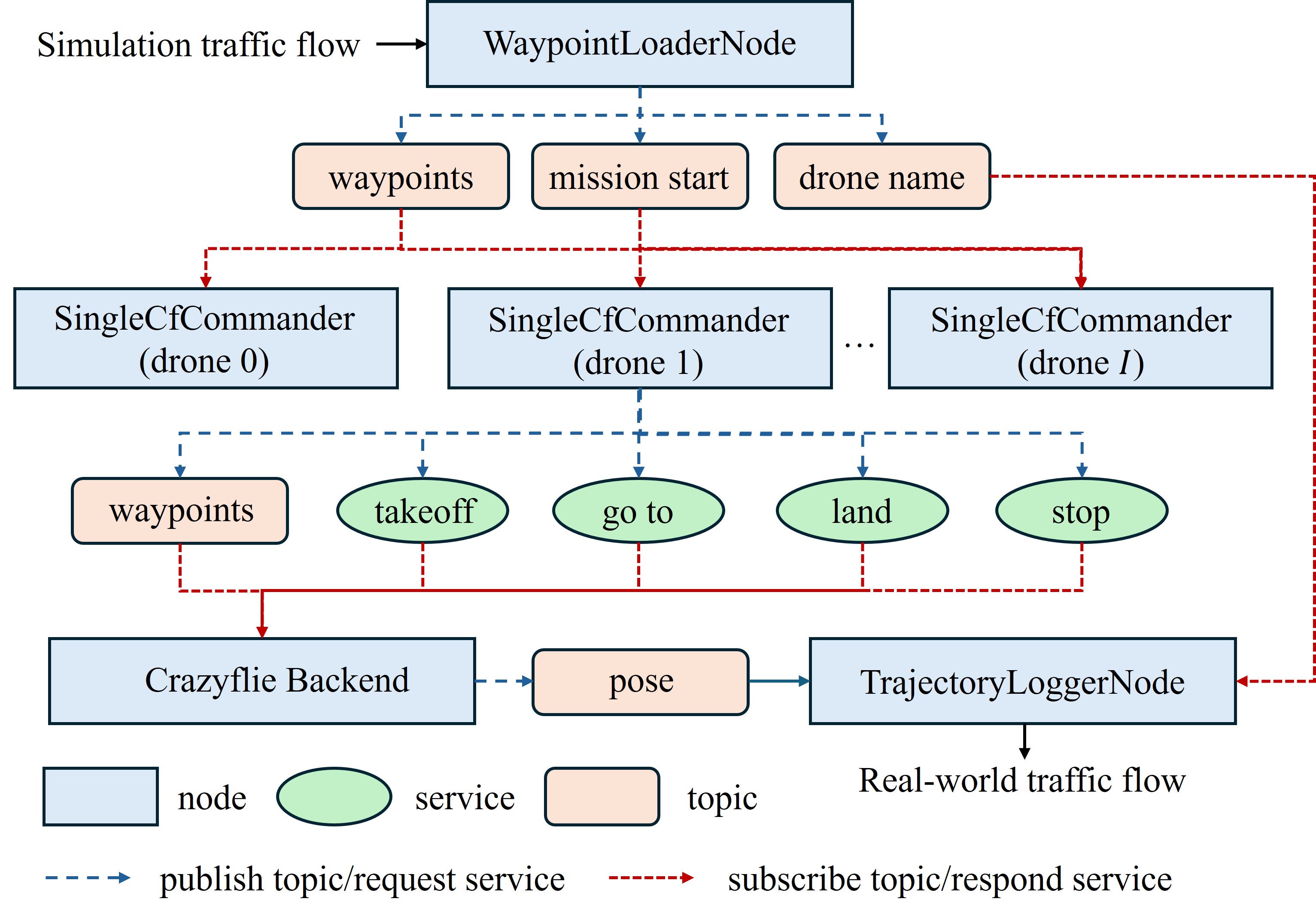}
    \caption{System architecture for drone swarm control using ROS~2 and Crazyswarm2.}
    \label{fig:ros2}
\end{figure}

Building on the ROS~2 middleware and the Crazyswarm2 framework, we developed a multi-process control architecture for coordinated trajectory tracking of multiple drones. This design ensures modularity, high-frequency command execution, and safe parallel operations across the drone swarm. The overall software architecture is illustrated in Figure~\ref{fig:ros2}. The system consists of four main types of nodes:
\begin{itemize}
    \item \textbf{WaypointLoaderNode}: loads predefined waypoint trajectories from CSV files and publishes them to the corresponding commanders via mission topics.
    \item \textbf{SingleCfCommander}: controls one drone by receiving its assigned waypoints and sending either high-frequency position commands or discrete movement requests to the backend. Each drone is associated with one independent commander process.
    \item \textbf{Crazyflie Backend}: serves as the communication interface between the ROS~2 network and the physical Crazyflie drones, providing services for takeoff, movement, and landing, as well as pose feedback.
    \item \textbf{TrajectoryLoggerNode}: subscribes to all drone pose topics and records the flight trajectories for post-experiment evaluation and analysis.
\end{itemize}

Based on this structure, a multi-process implementation is adopted instead of a multi-threaded one to avoid potential deadlocks in concurrent callback execution, which are common in ROS~2 multi-threading. Moreover, running each drone controller in an independent process ensures real-time responsiveness, allowing all drones to receive high-frequency control commands simultaneously without interference or scheduling delays.

\subsection{UAM trajectory-based traffic flow dataset}
\label{sec:data}

Following the methodology introduced in Section~\ref{sec:method} and the physical experiment setup described in Section~\ref{sec:phy}, we collect a trajectory-based UAM traffic flow dataset, \textbf{UAMTra2Flow}. The dataset consists of two components: (i) simulation data and (ii) physical experiment data, representing drone trajectories generated by the simulation framework and collected from real-world physical experiments, respectively. The physical experiment component contains approximately 46 minutes of flight trajectories, corresponding to about 140,000 rows of CSV-formatted data. The dataset is publicly available at \url{https://github.com/CATS-Lab/UAM-FD}. \textbf{The source code will be released upon acceptance of this paper.}

In the UAMTra2Flow dataset, we collect physical experiment data for two control laws across three scenario types. For each control law, two safety spacing settings are considered: for the stop-and-yield strategy, $\hat{h} \in \{0.5, 0.6\}$\,m; for the circular-detour strategy, $\hat{h} \in \{0.6, 0.7\}$\,m. To ensure that the collected data cover both free-flow and congested traffic regimes and thus enable reliable FD construction, experiments are conducted with different drone fleet sizes, $I \in \{2, 4, 6, 8\}$, for each configuration. In the physical experiments, each configuration is run once for each drone number, resulting in a total of 48 traffic flow trajectory segments. To obtain a larger number of data samples, each configuration in the simulation is repeated four times, yielding 192 traffic flow trajectory segments in total. In addition, readers can generate further simulation data or collect additional physical experiment trajectories using the released codebase.

Due to battery limitations, each test run lasts approximately 60\,s, including 5\,s for takeoff, 50\,s of steady flight, and 3\,s for landing. Trajectories are recorded at a sampling interval of $\Delta t = 0.1$\,s, and the desired cruising speed is set to $\bar{v} = 0.5$\,m/s. To prevent drones from completing all assigned flights prematurely, which would bias traffic flow measurements, we assign a sufficiently large number of OD flights $N$ so that all drones remain active throughout the experiment duration.

All trajectory data are stored in CSV files, with the column definitions summarized in Table~\ref{tab:data}. In particular, the dataset records both the drone’s current position and the coordinates of its next destination, which are required for computing the effective distance used in the FD analysis.

\begin{table}[t]
    \centering
    \caption{Structure of the UAMTra2Flow dataset.}
    \label{tab:data}
    \begin{tabular}{lll}
        \toprule
        \textbf{Column Name} & \textbf{Description} & \textbf{Unit} \\
        \midrule
        \texttt{id}        & Drone ID & N/A \\
        \texttt{time}      & Timestamp ($\Delta t=0.1$\,s) & s \\
        \texttt{px}        & $x$-coordinate of drone position & m \\
        \texttt{py}        & $y$-coordinate of drone position & m \\
        \texttt{pz}        & $z$-coordinate of drone position & m \\
        \texttt{dest\_px}  & $x$-coordinate of current destination & m \\
        \texttt{dest\_py}  & $y$-coordinate of current destination & m \\
        \texttt{dest\_pz}  & $z$-coordinate of current destination & m \\
        \bottomrule
    \end{tabular}
\end{table}

\section{Experiment results and discussions}
\label{sec:experiments}

This section presents the construction and analysis of the FDs using both the simulation data and the physical experiment data collected in Section~\ref{sec:data}. The results obtained from the different control laws and different scenario types are compared to evaluate the traffic characteristics of UAM. Finally, we scale the FD parameters to realistic conditions.

\subsection{Calibration of the FDs}

Using the UAMTra2Flow dataset, we calibrate the FD following the procedure described in Algorithm~\ref{alg:fd}. Prior to computing density and flow, only the trajectories recorded during the steady-flight phase are considered. Therefore, the first 5 seconds and the last 1 second of the trajectories (i.e., the takeoff and landing segments) are removed from the FD calibration.

\begin{figure}
    \centering
    \includegraphics[width=1.0\linewidth]{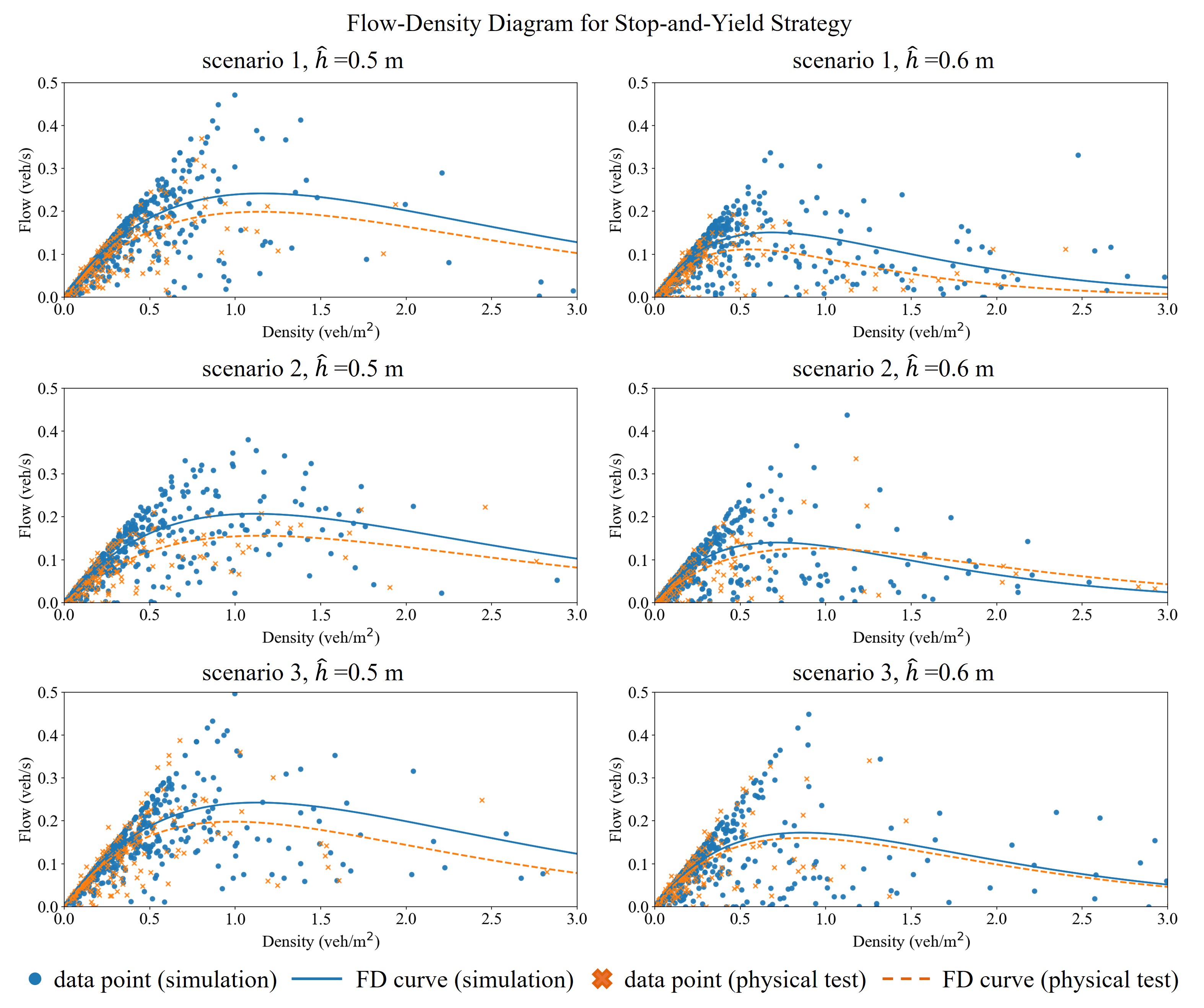}
    \caption{Flow-density diagrams for simulation and physical experiments for stop-and-yield strategy.}
    \label{fig:sim_phy_stop}
\end{figure}

\begin{figure}
    \centering
    \includegraphics[width=1.0\linewidth]{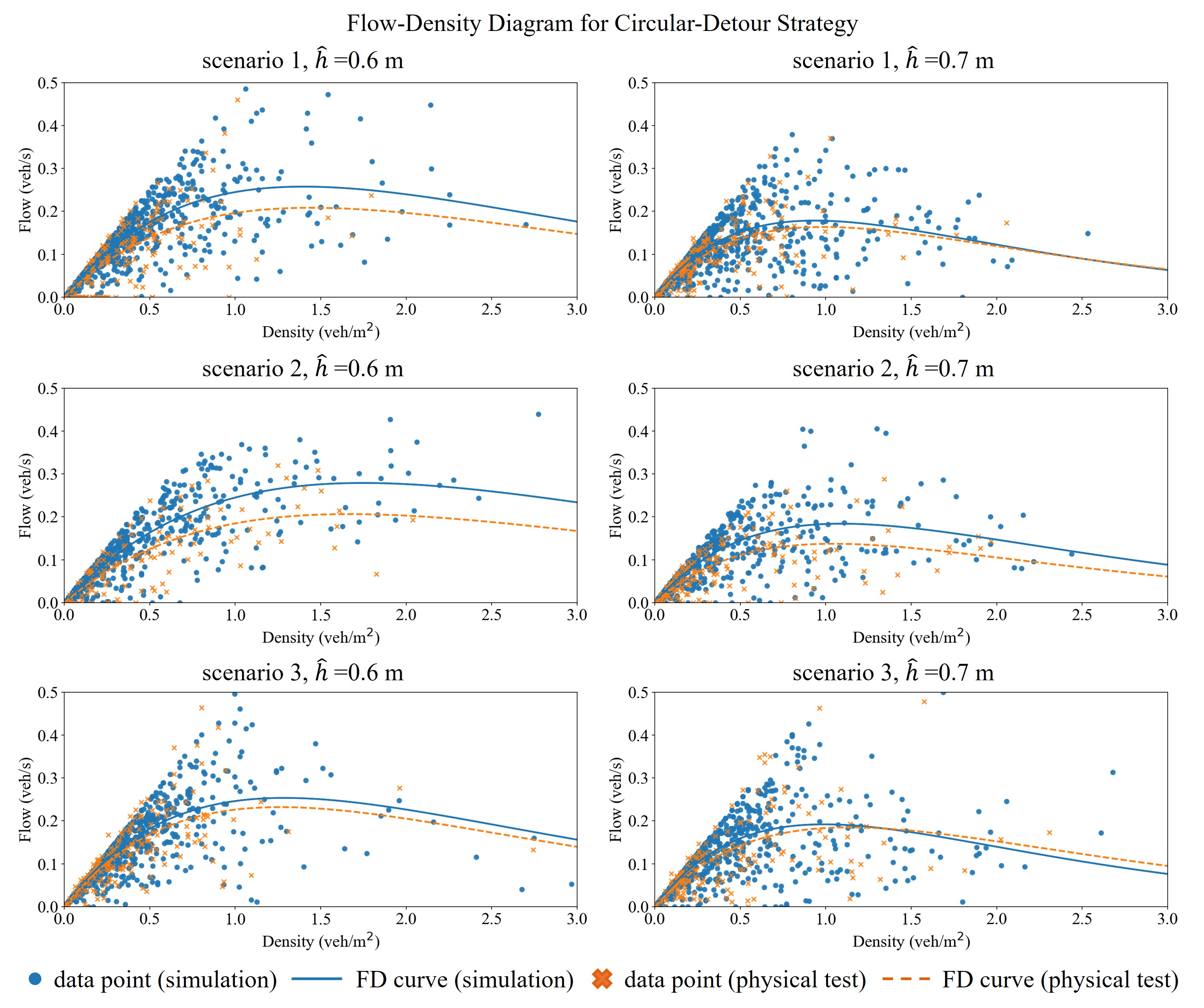}
    \caption{FD plots for simulation and physical experiments for circular-detour strategy.}
    \label{fig:sim_phy_detour}
\end{figure}

The calibrated FD parameters, together with the corresponding goodness-of-fit metrics including $R^2$ and RMSE, are reported in Tables~\ref{tab:FDcompare} and~\ref{tab:FDdetails}. Table~\ref{tab:FDcompare} compares the calibration results of different FD models, while the detailed parameters of the best-performing model are summarized in Table~\ref{tab:FDdetails}. The flow--density data points and the fitted FD curves for both simulation and physical experiments are illustrated in Figures~\ref{fig:sim_phy_stop}--\ref{fig:sim_phy_detour}.

\begin{table}[htbp]
    \centering
    \caption{Comparison of the calibration accuracy among different FD models. Here, \emph{stop} refers to the stop-and-yield strategy, and \emph{detour} refers to the circular-detour strategy. The unit of $\hat{h}$ is m, and the unit of RMSE is $\mathrm{m}\,\mathrm{s}^{-1}$.}
    \label{tab:FDcompare}
        \begin{tabular}{p{1.8cm}p{1.6cm}p{1.5cm}ccccccccc}
        \toprule
        \multirow{2}[4]{1.5cm}{Experiment type} & \multirow{2}[4]{1.4cm}{Scenario} & \multirow{2}[4]{1.3cm}{Control law} & \multirow{2}[4]{*}{$\hat{h}$} & \multicolumn{2}{c}{greenshields} & \multicolumn{2}{c}{greenberg} & \multicolumn{2}{c}{underwood} & \multicolumn{2}{c}{drake}  \\
        \cmidrule{5-12}
              &       &       &       & R2 & RMSE & R2 & RMSE & R2 & RMSE & R2 & RMSE \\
        \midrule
        \multirow{13}[2]{*}{simulation} & \multirow{4}[1]{*}{scenario 1} & \multirow{2}[1]{*}{stop} & \multicolumn{1}{l}{0.5} & 0.085 & 0.079 & 0.519 & 0.058 & \textbf{0.721} & \textbf{0.044} & 0.708 & 0.045 \\
              &       &       & \multicolumn{1}{l}{0.6} & 0.059 & 0.060 & 0.461 & 0.046 & \textbf{0.653} & \textbf{0.037} & 0.622 & 0.038 \\
              &       & \multirow{2}[0]{*}{detour} & \multicolumn{1}{l}{0.5} & 0.669 & 0.049 & 0.685 & \textbf{0.048} & \textbf{0.689} & \textbf{0.048} & 0.659 & 0.050 \\
              &       &       & \multicolumn{1}{l}{0.6} & 0.498 & 0.051 & 0.573 & \textbf{0.047} & \textbf{0.579} & \textbf{0.047} & 0.530 & 0.049 \\
              & \multirow{4}[0]{*}{scenario 2} & \multirow{2}[0]{*}{stop} & \multicolumn{1}{l}{0.6} & 0.745 & 0.040 & 0.783 & 0.037 & \textbf{0.794} & \textbf{0.036} & 0.783 & 0.037 \\
              &       &       & \multicolumn{1}{l}{0.7} & 0.420 & 0.045 & 0.591 & 0.038 & \textbf{0.652} & \textbf{0.035} & 0.639 & 0.036 \\
              &       & \multirow{2}[0]{*}{detour} & \multicolumn{1}{l}{0.6} & 0.799 & 0.042 & 0.815 & \textbf{0.040} & \textbf{0.817} & \textbf{0.040} & 0.793 & 0.042 \\
              &       &       & \multicolumn{1}{l}{0.7} & 0.699 & 0.041 & 0.712 & \textbf{0.040} & \textbf{0.716} & \textbf{0.040} & 0.697 & 0.041 \\
              & \multirow{4}[0]{*}{scenario 3} & \multirow{2}[0]{*}{stop} & \multicolumn{1}{l}{0.5} & 0.507 & 0.062 & 0.672 & 0.051 & \textbf{0.708} & \textbf{0.048} & 0.706 & 0.048 \\
              &       &       & \multicolumn{1}{l}{0.6} & 0.247 & 0.062 & 0.482 & 0.051 & \textbf{0.558} & \textbf{0.047} & 0.538 & 0.048 \\
              &       & \multirow{2}[0]{*}{detour} & \multicolumn{1}{l}{0.5} & 0.708 & 0.048 & 0.702 & 0.048 & 0.723 & 0.047 & \textbf{0.730} & \textbf{0.046} \\
              &       &       & \multicolumn{1}{l}{0.6} & 0.467 & 0.059 & 0.567 & 0.053 & \textbf{0.575} & \textbf{0.053} & 0.537 & 0.055 \\
              & \multicolumn{3}{c}{average} & 0.492 & 0.053 & 0.630 & 0.046 & \textbf{0.682} & \textbf{0.043} & 0.662 & 0.045 \\
        \midrule
        \multirow{13}[2]{*}{physics} & \multirow{4}[1]{*}{scenario 1} & \multirow{2}[1]{*}{stop} & \multicolumn{1}{l}{0.5} & 0.642 & 0.042 & 0.680 & 0.040 & \textbf{0.690} & \textbf{0.039} & 0.678 & 0.040 \\
              &       &       & \multicolumn{1}{l}{0.6} & 0.097 & 0.047 & 0.423 & 0.037 & \textbf{0.565} & \textbf{0.032} & 0.555 & 0.033 \\
              &       & \multirow{2}[0]{*}{detour} & \multicolumn{1}{l}{0.5} & \textbf{0.562} & 0.053 & 0.552 & 0.053 & \textbf{0.562} & \textbf{0.053} & 0.557 & 0.053 \\
              &       &       & \multicolumn{1}{l}{0.6} & 0.538 & 0.045 & 0.573 & 0.043 & \textbf{0.581} & \textbf{0.043} & 0.548 & 0.044 \\
              & \multirow{4}[0]{*}{scenario 2} & \multirow{2}[0]{*}{stop} & \multicolumn{1}{l}{0.6} & 0.662 & 0.034 & \textbf{0.725} & \textbf{0.030} & 0.722 & \textbf{0.030} & 0.669 & 0.033 \\
              &       &       & \multicolumn{1}{l}{0.7} & 0.285 & 0.044 & 0.532 & 0.035 & 0.649 & 0.031 & \textbf{0.672} & \textbf{0.030} \\
              &       & \multirow{2}[0]{*}{detour} & \multicolumn{1}{l}{0.6} & 0.770 & 0.037 & 0.756 & 0.038 & 0.766 & 0.038 & \textbf{0.770} & \textbf{0.037} \\
              &       &       & \multicolumn{1}{l}{0.7} & 0.572 & 0.039 & 0.592 & \textbf{0.038} & \textbf{0.598} & \textbf{0.038} & 0.571 & 0.039 \\
              & \multirow{4}[0]{*}{scenario 3} & \multirow{2}[0]{*}{stop} & \multicolumn{1}{l}{0.5} & 0.442 & 0.056 & 0.540 & 0.051 & \textbf{0.570} & \textbf{0.049} & 0.555 & 0.050 \\
              &       &       & \multicolumn{1}{l}{0.6} & 0.220 & 0.058 & 0.501 & 0.046 & \textbf{0.600} & \textbf{0.041} & 0.586 & 0.042 \\
              &       & \multirow{2}[0]{*}{detour} & \multicolumn{1}{l}{0.5} & 0.646 & 0.048 & 0.653 & 0.047 & \textbf{0.669} & \textbf{0.046} & 0.657 & 0.047 \\
              &       &       & \multicolumn{1}{l}{0.6} & 0.517 & 0.059 & 0.523 & 0.059 & 0.535 & \textbf{0.058} & \textbf{0.536} & \textbf{0.058} \\
              & \multicolumn{3}{c}{average} & 0.496 & 0.047 & 0.587 & 0.043 & \textbf{0.626} & \textbf{0.041} & 0.613 & 0.042 \\
        \bottomrule
    \end{tabular}
\end{table}

Table~\ref{tab:FDcompare} compares the calibration results of five FD models, including the Greenshields model \citep{greenshields1935study}, the Greenberg model \citep{greenberg1959analysis}, the Underwood model \citep{underwood1960speed}, and the Drake model \citep{drake1967statistical}.
From the calibration results, the Underwood model achieve higher $R^2$ values and lower RMSE values than the other models, indicating better fitting performance for UAM traffic data. The Drake and Greenberg models show moderate performance, while the Greenshields model consistently exhibits lower calibration accuracy and appears less suitable for UAM applications. Overall, the calibration accuracy obtained from physical experiments is slightly lower than that from simulation, which is expected due to real-world disturbances and modeling uncertainties. Based on these results, the following discussion of FD characteristics is conducted using the Underwood model.

\begin{table}[t]
    \small
    \caption{Fitted parameters and calibration results of Underwood's model. Here, \emph{stop} refers to the stop-and-yield strategy, and \emph{detour} refers to the circular-detour strategy.}
    \label{tab:FDdetails}
    \centering
    \begin{tabular}{p{1.8cm}p{1.5cm}p{2cm}cccccc}
    \toprule
    Scenario type & Control law & Experiment type & $\hat{h}$ (m) & $q_{\max}$ ($\mathrm{m}^{-1}\mathrm{s}^{-1}$) & $k_{\mathrm{c}}$ ($\mathrm{m}^{-2}$)  & $v_{\mathrm{f}}$ ($\mathrm{m}\,\mathrm{s}^{-1}$) & $R^2$ & RMSE ($\mathrm{m}\,\mathrm{s}^{-1}$)\\
    \midrule
    \multirow{8}[1]{*}{scenario 1} & \multirow{4}[1]{*}{stop} & \multirow{2}[1]{*}{simulation} & 0.5   & 0.236 & 1.173 & 0.546 & 0.721 & 0.044 \\
          &       &       & 0.6   & 0.145 & 0.660 & 0.599 & 0.653 & 0.037 \\
          &       & \multirow{2}[0]{*}{physics} & 0.5   & 0.188 & 1.073 & 0.476 & 0.690 & 0.039 \\
          &       &       & 0.6   & 0.109 & 0.533 & 0.557 & 0.565 & 0.032 \\
          & \multirow{4}[0]{*}{detour} & \multirow{2}[0]{*}{simulation} & 0.6   & 0.250 & 1.382 & 0.491 & 0.689 & 0.048 \\
          &       &       & 0.7   & 0.175 & 0.933 & 0.510 & 0.579 & 0.047 \\
          &       & \multirow{2}[0]{*}{physics} & 0.6   & 0.217 & 1.640 & 0.359 & 0.562 & 0.053 \\
          &       &       & 0.7   & 0.151 & 0.874 & 0.468 & 0.581 & 0.043 \\
    \multirow{8}[0]{*}{scenario 2} & \multirow{4}[0]{*}{stop} & \multirow{2}[0]{*}{simulation} & 0.5   & 0.201 & 1.095 & 0.499 & 0.794 & 0.036 \\
          &       &       & 0.6   & 0.132 & 0.665 & 0.537 & 0.652 & 0.035 \\
          &       & \multirow{2}[0]{*}{physics} & 0.5   & 0.152 & 1.140 & 0.362 & 0.722 & 0.030 \\
          &       &       & 0.6   & 0.131 & 0.909 & 0.391 & 0.649 & 0.031 \\
          & \multirow{4}[0]{*}{detour} & \multirow{2}[0]{*}{simulation} & 0.6   & 0.276 & 1.777 & 0.423 & 0.817 & 0.040 \\
          &       &       & 0.7   & 0.177 & 1.081 & 0.445 & 0.716 & 0.040 \\
          &       & \multirow{2}[0]{*}{physics} & 0.6   & 0.202 & 1.657 & 0.332 & 0.766 & 0.038 \\
          &       &       & 0.7   & 0.130 & 1.003 & 0.352 & 0.598 & 0.038 \\
    \multirow{8}[1]{*}{scenario 3} & \multirow{4}[0]{*}{stop} & \multirow{2}[0]{*}{simulation} & 0.5   & 0.230 & 1.066 & 0.586 & 0.708 & 0.048 \\
          &       &       & 0.6   & 0.154 & 0.733 & 0.571 & 0.558 & 0.047 \\
          &       & \multirow{2}[0]{*}{physics} & 0.5   & 0.179 & 0.868 & 0.561 & 0.570 & 0.049 \\
          &       &       & 0.6   & 0.148 & 0.771 & 0.524 & 0.600 & 0.041 \\
          & \multirow{4}[1]{*}{detour} & \multirow{2}[0]{*}{simulation} & 0.6   & 0.246 & 1.245 & 0.537 & 0.723 & 0.047 \\
          &       &       & 0.7   & 0.186 & 1.026 & 0.494 & 0.575 & 0.053 \\
          &       & \multirow{2}[1]{*}{physics} & 0.6   & 0.218 & 1.189 & 0.499 & 0.669 & 0.046 \\
          &       &       & 0.7   & 0.182 & 1.106 & 0.448 & 0.535 & 0.058 \\

    \bottomrule
    \end{tabular}
\end{table}

Table~\ref{tab:FDdetails} shows the detailed parameters for the calibrated Underwood model. In this table, $q_{\max}$ denotes the maximum flow rate (unit: $\text{m}^{-1}\text{s}^{-1}$), $k_{\mathrm{c}}$ (unit: $\text{m}^{-2}$) represents the critical density, and $v_{\mathrm{f}}$ (unit: $\text{m}\,\text{s}^{-1}$) is the free-flow speed. Based on the $R^2$ and RMSE values, the regression models exhibit overall strong performance when compared with related studies \citep{shi2021constructing}. For the simulation data, all calibrated models achieve relatively high fitting accuracy. The average $R^2$ value is 0.682, and the minimum $R^2$ value is 0.575. The average RMSE is 0.043, and the maximum RMSE is 0.053. In comparison, the calibration results obtained from physical experiments are slightly less accurate. The average $R^2$ value is 0.626, and the minimum $R^2$ value is 0.535. The average RMSE is 0.041, and the maximum RMSE is 0.058. These results suggest that traditional FD models originally developed for ground transportation remain effective for characterizing the macroscopic traffic behavior of UAM systems.

Besides, for both Table~\ref{tab:FDdetails} and Figure~\ref{fig:sim_phy_stop}-\ref{fig:sim_phy_detour}, we can also observe that there are some differences between the FD curves obtained from simulation and physical experiment. Specifically, the FD curves derived from physical experiments typically exhibit lower free-flow speeds and smaller maximum flows compared to those obtained from simulations. Furthermore, the data from physical experiments fit parametric models less well in general. 
These differences are likely due to inherent characteristics of real-world UAV systems that are not fully captured in idealized simulations. In practice, UAV operations involve finite control bandwidth, actuator limits, and non-negligible sensing and communication delays. These factors can reduce the realized speed relative to the commanded speed and make collision-avoidance behavior more conservative, especially under high-density conditions. In addition, environmental disturbances such as aerodynamic drag and multi-vehicle airflow interactions may further affect the effective traffic dynamics. As a result, the calibrated FD obtained from physical experiments exhibits lower free-flow speeds and smaller maximum flows. This phenomenon is consistent with the classical discrepancy between theoretical and empirical FDs observed in ground vehicular traffic \citep{bramich2022fitting}, which highlights the importance of validating simulation-based conclusions under realistic operational conditions.

\subsection{Comparison between different control laws}

\begin{figure}
    \centering
    \includegraphics[width=1.\linewidth]{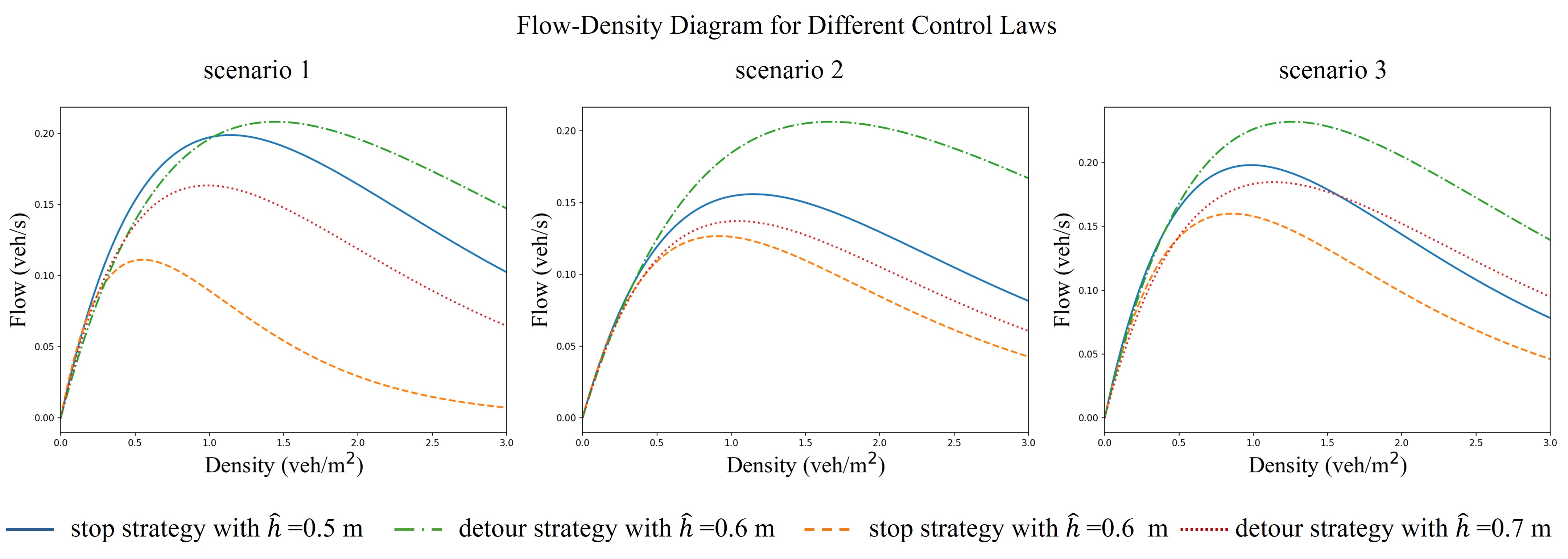}
    \caption{Flow-density diagrams for the stop-and-yield strategy and the circular-detour strategy with different safety spacing.}
    \label{fig:controlFD}
\end{figure}

To compare the characteristics of different control laws, we plot the FD curves obtained under the stop-and-yield strategy and the circular-detour strategy in Figure~\ref{fig:controlFD}. When comparing the same control law under different safety spacing settings, a clear pattern emerges: larger safety spacing consistently leads to a lower maximum flow and critical density. This observation indicates that traffic capacity, defined as the maximum achievable flow rate, decreases as the safety spacing increases, which is consistent with well-established findings in ground transportation systems.

When comparing different control laws under the same safety spacing setting, the circular-detour strategy consistently produces a higher maximum flow rate, indicating superior traffic capacity. A plausible explanation is that the circular-detour strategy allows drones involved in potential conflicts to maintain continuous motion by dynamically adjusting their directions, thereby preventing the formation of localized blockages. In contrast, under the stop-and-yield strategy, once multiple drones accumulate near a conflict region, low-priority drones may remain stationary for extended periods, which amplifies congestion and substantially reduces throughput.

Finally, we observe that both control laws exhibit similar free-flow speeds across different safety spacing settings. This overall consistency in free-flow speed supports the effectiveness of the proposed effective-distance measurement in Section \ref{sec:measure}, which preserves accurate flow estimation in the uncongested regime.

\subsection{Comparison between different application scenarios}

\begin{figure}
    \centering
    \includegraphics[width=1\linewidth]{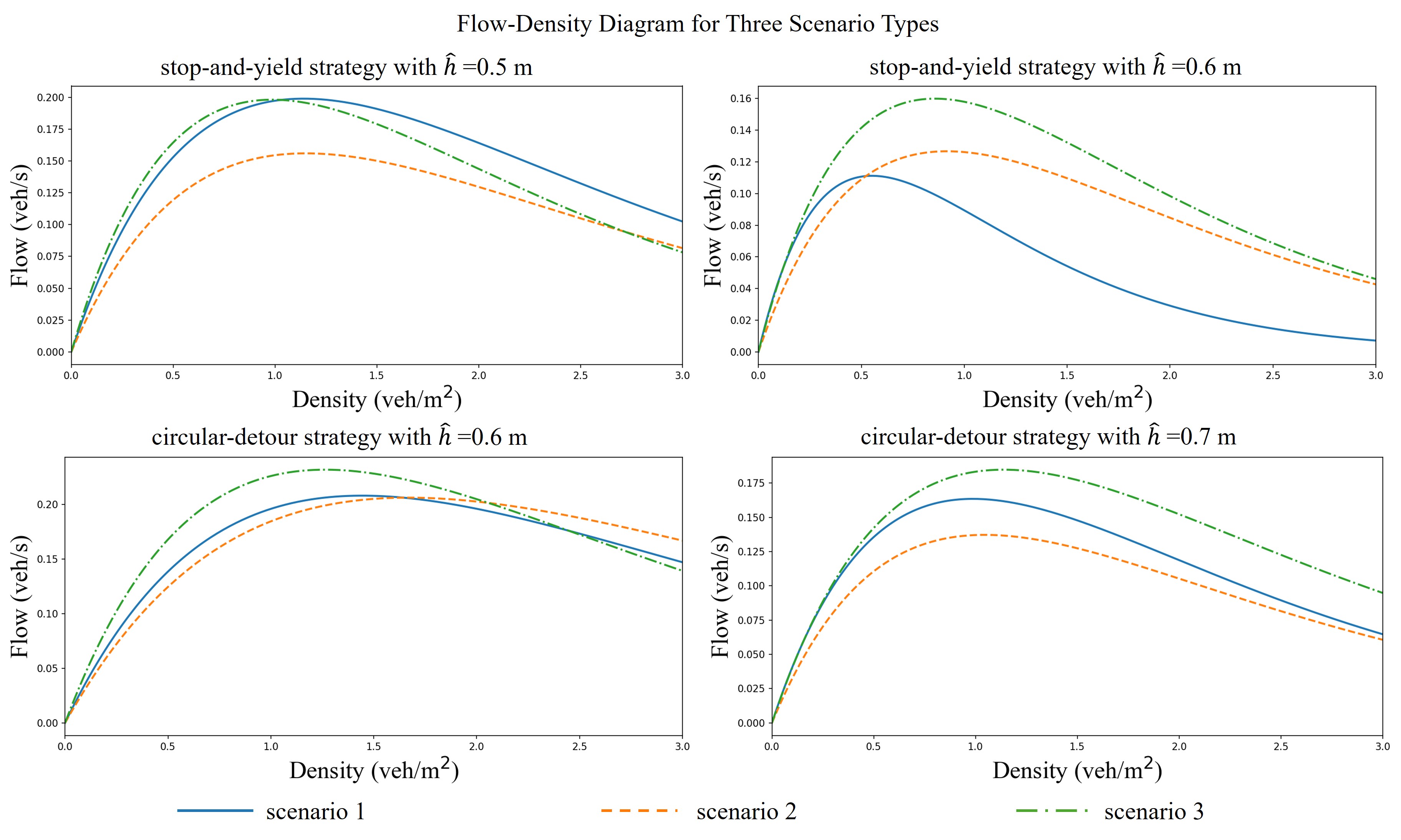}
    \caption{FD plots for comparison for three scenario types.}
    \label{fig:scenario}
\end{figure}

Figure~\ref{fig:scenario} compares the FD curves across different scenario types. Overall, no single scenario consistently dominates the others in terms of congestion level, as the relative traffic performance varies depending on the control law and safety spacing settings. Nevertheless, a general trend can be observed: Scenario~3 tends to exhibit less severe congestion compared with the other two scenario types. A plausible explanation is that Scenario~3 represents station-to-station operations between fixed locations, where drone trajectories are more structured and recurrent, effectively forming lane-like patterns in the airspace. In contrast, in Scenarios~1 and~2, where origin–destination pairs are generated randomly, drone movements are more disordered, which increases interaction complexity and makes congestion more likely to occur.

\subsection{Scaling to real-world UAM}
\label{subsec:scaling}

To bridge the gap between reduced-scale physical experiments and real-world UAM traffic, we attempt to scale the calibrated FD parameters to realistic operational settings. We assume that the FD is mainly influenced by the characteristic cruising speed and the effective spatial occupancy induced by vehicle size and safety spacing. Under this assumption, traffic density scales with the inverse of the occupied area, while traffic flow scales linearly with the characteristic speed.

Let $\eta$ and $\eta'$ denote the characteristic length of a drone in the reduced-scale testbed and in the realistic operational setting, respectively, and let $\bar{v}'$ denote the desired cruising speed in the realistic setting. Define the scaling factors for size and speed as $\delta_{\eta}:= \frac{\eta'}{\eta}$ and $\delta_{v}:= \frac{\bar{v}'}{\bar{v}}$. In our FD, density is measured in $\mathrm{veh}/\mathrm{m}^2$ and flow in $\mathrm{veh}/(\mathrm{m}\cdot\mathrm{s})$, so the scaled FD parameters can be expressed as
\begin{align}
    v'_{\mathrm{f}} = v_{\mathrm{f}}\,\delta_{v},
    \qquad
    k'_{\mathrm{c}} = \frac{k_{\mathrm{c}}}{\delta_{\eta}^{2}},
    \qquad
    q'_{\max} = q_{\max}\,\frac{\delta_{v}}{\delta_{\eta}^{2}},
    \label{eq:scale_fd_params}
\end{align}

In the experiments, the size of the Crazyflie drone is approximately $10\,\mathrm{cm}\times10\,\mathrm{cm}$, and thus we take $\eta = 10$~cm. Referring to the specifications of a representative large-scale UAM vehicle, the DJI FlyCart~100~\citep{DJI_FlyCart100_specs}, we set the realistic drone size to $\eta' = 2$~m and cruising speed $\bar{v}' = 10$~m/s. Using these parameters, the calibrated FD results in Table~\ref{tab:FDdetails} are scaled to realistic UAM settings, as reported in Table~\ref{tab:scale}. Here, $\hat{h}'$ denotes the scaled safety spacing, with units of m. For ease of comparison with real-world data, all other FD parameters are converted to a km--h unit system.

\begin{table}[htbp]
    \centering
    \caption{FD parameters scaled to realistic UAM traffic.}
    \begin{tabular}{lllll}
    \toprule
    Scenario type& Control law & $\hat{h}'$ (m) & $q'_{\max}$ ($\mathrm{km}^{-1}\mathrm{h}^{-1}$) & $k'_{\mathrm{c}}$ ($\text{km}^{-2}$) \\
    \midrule
    \multirow{4}[1]{*}{scenario 1} & \multirow{2}[0]{*}{stop} & 10    & 38596.6 & 2603.5 \\
          &       & 12    & 18296.9 & 1462.5 \\
          & \multirow{2}[1]{*}{detour} & 12    & 31154.1 & 2298.2 \\
          &       & 14    & 30716.8 & 1992.3 \\
    \midrule
    \multirow{4}[2]{*}{scenario 2} & \multirow{2}[1]{*}{stop} & 10    & 35547.9 & 3310.8 \\
          &       & 12    & 19467.5 & 1608.9 \\
          & \multirow{2}[1]{*}{detour} & 12    & 40998.5 & 3998.4 \\
          &       & 14    & 28737.0 & 2566.2 \\
    \midrule
    \multirow{4}[2]{*}{scenario 3} & \multirow{2}[1]{*}{stop} & 10    & 32218.3 & 1932.9 \\
          &       & 12    & 24222.6 & 1448.7 \\
          & \multirow{2}[1]{*}{detour} & 12    & 37644.8 & 2283.7 \\
          &       & 14    & 35090.3 & 2568.1 \\
    \bottomrule
    \end{tabular}%
  \label{tab:scale}%
\end{table}%

\section{Conclusion and future research}
\label{sec:conclusion}

This study presents a general framework for constructing the FD of UAM traffic using both simulation and physical experiments. From the theoretical perspective, the proposed framework integrates UAM traffic definitions, control law design, and traffic flow theory to generate UAM traffic and characterize the relationship between UAM flow and density under diverse operational conditions. From the experimental perspective, a reduced-scale UAM testbed based on Crazyflie drones and the ROS~2 system is established to enable accurate drone control and reliable trajectory data collection, thereby supporting real-world UAM traffic analysis. Based on the proposed framework, the UAMTra2Flow dataset is collected, which can facilitate future studies on UAM traffic modeling and management. Analysis of the FDs derived from the UAMTra2Flow dataset confirms that classical traffic flow patterns developed for ground transportation are largely applicable to UAM systems, while also revealing systematic deviations from simulation-based results. These findings underscore the importance of incorporating physical experiments in UAM traffic studies to better capture real-world system dynamics. Moreover, the results demonstrate that different control laws have distinct impacts on UAM traffic performance, motivating the design of appropriate control strategies for efficient and reliable UAM traffic.

Although the proposed framework has been preliminarily validated, several limitations remain. First, the number of experimental runs is limited, which constrains the extent to which the effects of control parameters and modeling assumptions can be systematically analyzed. Second, despite incorporating both theoretical analysis and physical experiments, the current study still relies on certain simplifications. For example, the control laws are considered with fixed safety spacing, whereas in reality, the safety spacing might be different according to the current speed. These simplifications may result in the current experiment not capturing the full diversity and variability of future UAM systems. Finally, the unique characteristics of UAM traffic may require dedicated FD calibration methods to further improve accuracy, especially to mitigate the gap between empirical and theoretical FDs.
Future work can be conducted along several directions. First, the number of experimental scenarios can be expanded to support more comprehensive analyses. Second, more realistic operational scenarios and control laws can be designed to better reflect real-world UAM traffic. Finally, advanced filtering or data-selection methods adapted from the traditional FD research \citep{bramich2022fitting} for flow--density observations can be explored to improve FD calibration accuracy. These efforts will further enhance the applicability of FD-based approaches in guiding UAM traffic management and infrastructure development.

\bibliographystyle{informs2014}
\bibliography{reference.bib}

\end{document}